# Overview of CHIP 2025 Shared Task 2: Discharge Medication Recommendation for Metabolic Diseases Based on Chinese Electronic Health Records

Juntao Li[1], Haobin Yuan[1], Ling Luo[1,(✉)], Tengxiao Lv[1], Yan Jiang[2], Fan Wang[2], Ping Zhang[2], Huiyi Lv[2], Jian Wang[1], Yuanyuan Sun[1] and Hongfei Lin[1]

[1] Dalian University of Technology, Dalian 116024, Liaoning, China
[2] The Second Affiliated Hospital of Dalian Medical University, Dalian 116023, Liaoning, China
`lingluo@dlut.edu.cn`

**Abstract.** Discharge medication recommendation plays a critical role in ensuring treatment continuity, preventing readmission, and improving long-term management for patients with chronic metabolic diseases. This paper present an overview of the CHIP 2025 Shared Task 2 competition, which aimed to develop state-of-the-art approaches for automatically recommending appropriate discharge medications using real-world Chinese EHR data. For this task, we constructed CDrugRed, a high-quality dataset consisting of 5,894 de-identified hospitalization records from 3,190 patients in China. This task is challenging due to multi-label nature of medication recommendation, heterogeneous clinical text, and patient-specific variability in treatment plans. A total of 526 teams registered, with 167 and 95 teams submitting valid results to the Phase A and Phase B leaderboards, respectively. The top-performing team achieved the highest overall performance on the final test set, with a Jaccard score of 0.5102, F1 score of 0.6267, demonstrating the potential of advanced large language model (LLM)-based ensemble systems. These results highlight both the promise and remaining challenges of applying LLMs to medication recommendation in Chinese EHRs. The post-evaluation phase remains open at https://tianchi.aliyun.com/competition/entrance/532411/.

## 1 Introduction

With the global incidence of chronic metabolic diseases—such as diabetes, hypertension, and fatty liver disease—continuing to rise, their treatment and management have become increasingly complex. Patients with metabolic disorders often suffer from multiple comorbidities, require long-term pharmacological treatment, and exhibit significant individual variability. The selection of medications appropriate medications after hospital discharge plays a critical role in ensuring long-term disease control, preventing readmission, and improving treatment safety. During hospitalization, rich diagnostic, laboratory, and treatment information is continuously recorded in electronic health records (EHRs). These records provide valuable real-world data for developing precise and intelligent discharge medication recommendation systems that can assist clinicians in optimizing personalized treatment strategies [1-3].

In recent years, drug recommendation has received increasing research attention, with most studies primarily focusing on English-language datasets such as MIMIC-III [4] and MIMIC-IV [5]. Despite the growing research interest, there remains a lack of standardized shared tasks for drug recommendation based on real-world EHR data. In particular, research and evaluation efforts based on Chinese clinical data are still in their early stages. Publicly available benchmark datasets and standardized evaluation criteria are essential to ensure fair comparisons across different approaches. Therefore, we organized the discharge medication recommendation task for metabolic diseases based on Chinese EHRs at the China Health Information Processing Conference (CHIP), aiming to promote research on intelligent medication recommendation in real-world Chinese healthcare settings [6]. As shown in **Fig. 1**, the task involves inputting de-identified inpatient Chinese EHR texts into the models to automatically recommend drugs selected from a predefined candidate list.

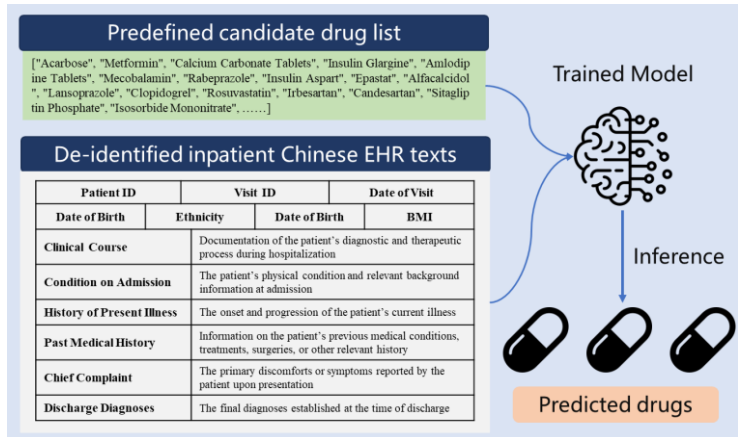

**Fig. 1.** Overview of the Discharge Medication Recommendation Task

The task's evaluation competition was hosted on the Tianchi platform[1] and consisted of two phases: Phase A (development set) for model training and Phase B (test set) for final evaluation. The competition attracted active participation from universities, enterprises, and research institutions. A total of 526 teams registered, among which 167 teams submitted results for the Phase A leaderboard, and 95 teams participated in the Phase B evaluation. The results revealed numerous promising approaches for Chinese EHR-based drug recommendation. The top-ranked team achieved the best performance, with a Jaccard score of 0.5102 and an F1 score of 0.6267, demonstrating the feasibility and potential of applying advanced machine learning and natural language processing methods to discharge medication recommendation tasks.

---

[1]    https://tianchi.aliyun.com/competition/entrance/532411/information

To further advance research in the field, a post-evaluation phase was launched after the competition to encourage continued exploration using the dataset. In this paper, the following sections first review related work, then provide a detailed description of the dataset construction, task definition and evaluation metric. Subsequently, we analyze the key methods adopted by the top-performing teams and their results, and finally present a summary and outlook for this evaluation.

## 2    Related Work

To promote the development of artificial intelligence technologies in the biomedical domain, a growing number of challenge evaluations have been organized, such as BioCreative[2], China Conference on Knowledge Graph and Semantic Computing (CCKS) [7], China National Conference on Computational Linguistics (CCL), and CHIP. Although many of these evaluations are held annually and cover a wide range of biomedical topics, tasks specifically targeting medication recommendation remain relatively scarce. Recently, Task 9 of CCL25-Eval[3] has focused on the core challenges of syndrome differentiation and herbal prescription generation in Traditional Chinese Medicine. Given clinical symptom descriptions as input, participants are required to generate personalized herbal prescriptions. In contrast, our task centers on modern medicine, targeting discharge medication recommendation for metabolic diseases based on real electronic medical records. Moreover, our evaluation is supported by a larger, high-quality dataset, enabling more comprehensive benchmarking and facilitating the development of robust, data-driven medication recommendation models.

With the continuous development of large language models (LLMs) [8-12], their applications in the healthcare have expanded rapidly, leading to the emergence of numerous medical LLMs [13-15]. These models are typically trained on medical corpora and have achieved strong performance across various benchmark tasks [16-18]. For example, Med-PaLM 2 [19] integrates large-scale, high-quality medical data with multi-stage instruction fine-tuning to enhance capabilities in medical question answering, clinical reasoning, medical knowledge retrieval, and professional text generation. In the Chinese medical context, LLMs such as HuatuoGPT-o1 [20] have been developed for complex medical reasoning tasks. HuatuoGPT-o1 is trained on a dataset of 40,000 verifiable medical questions and employing a medical verifier to evaluate output quality, thereby enabling quantifiable optimization of reasoning pathways.

In recent years, a growing number of open-source medical datasets have also been released [21-24], most of which serve as benchmarks for evaluating the performance of medical LLMs. Research on discharge medication recommendation heavily relies on open EHR and clinical text data for model training and validation [25-27]. Among these datasets [4,5,28], the MIMIC series [4,5] from Beth Israel Hospital is one of the most widely used, playing a crucial role in studies on diagnosis prediction, prognosis modeling, and medication recommendation. However, most existing datasets for drug recommendation are based on English-language data, and Chinese datasets remain scarce. The DAILMED dataset [29], which adopts a dialogue-based paradigm for medication recommendation, is among of the few available Chinese datasets in this field. Therefore, we introduce CDrugRed [30], the Chinese discharge medication recommendation dataset focusing on metabolic diseases. This dataset provides a valuable foundation for developing and evaluating intelligent medication recommendation systems tailored to Chinese clinical practice.

---

[2] https://www.ncbi.nlm.nih.gov/research/bionlp/biocreative
[3] https://tianchi.aliyun.com/competition/entrance/532301

## 3      Material and Methods

### 3.1      Task Definition

In this shared task, the systems of participating teams are required to predict a set of discharge medications for patients with metabolic diseases based on de-identified Chinese EHR texts. The recommended medications must be selected from a predefined candidate drug list. The input to the model consists of the patient's inpatient records, and the output is a patient-specific list of medications that would be appropriately prescribed at discharge. Participants were encouraged to explore diverse technical solutions, including traditional machine learning, deep learning, pretrained language models, or retrieval-augmented generation methods. To ensure fairness and reproducibility, the task imposed certain constraints: 1) The total model size was limited to 10 billion parameters; 2) The use of external resources or knowledge bases was permitted only if their sources were explicitly declared. These standardized requirements aimed to promote methodological innovation while maintaining the integrity and comparability of system performance across participating teams.

### 3.2      Dataset Description

The evaluation was conducted using the CDrugRed dataset [30], a new Chinese discharge medication recommendation dataset focused on metabolic diseases. This dataset was developed to facilitate research on intelligent and interpretable clinical medication recommendation. CDrugRed is constructed from real electronic medical records collected at a top-tier tertiary hospital in China, covering 5,894 de-identified hospitalization records from 3,190 patients spanning the years 2013 and 2023. The dataset includes 651 candidate drugs and provides rich clinical information, including basic demographic data, admission conditions, inpatient clinical course, laboratory and examination results, past medical history, discharge diagnoses, and actual discharge medications prescribed by physicians. An example is shown in **Table 1**. To protect patient privacy and ensure data quality, the dataset underwent a multi-stage processing pipeline involving large-model-assisted de-identification, standardized drug-name normalization, and manual review to ensure data security and consistency with clinical terminology.

Table 1. Example data format

| Items | Example |
| --- | --- |
| Patient ID | 2 |
| Visit ID | "2-1" |
| Sex | "Female" |
| Date of Birth | "1940-12" |
| Ethnicity | "Han Chinese" |
| BMI | 27.3 |
| Date of Visit | "2015-03" |
| Clinical Course | "Outpatient urine test: ketone bodies: +, leukocytes: 250/μl; microscopic examination: 7–9/HP. Upon admission: urinalysis: leukocytes +++, urine leukocytes 62.25/HP (↑). After oral rehydration, repeat urinalysis showed leukocytes and ketones (-). Steamed bun meal test results reported: glycated hemoglobin......" |
| Condition on Admission | "The patient was admitted with the chief complaint of polydipsia, polyuria, and polyphagia for 5 years, dysuria with poor glycemic control for 2 months. Key physical examination findings: T 36.6 °C, P 76 bpm, R 22 breaths/min, BP 160/80 mmHg......" |
| History of Present Illness | "Five years prior, the patient developed polydipsia, polyuria, and polyphagia without obvious precipitating factors, and presented to ** Hospital. Fasting blood glucose was 16.7 mmol/L. The patient was treated with metformin, repaglinide, and acarbose for glycemic control......" |
| Past Medical History | "No history of coronary heart disease. Denies history of infectious diseases including hepatitis, tuberculosis, and malaria. No history of food or drug allergy. No history of trauma, surgery, or blood transfusion. Immunization history is unknown." |
| Chief Complaint | "Polydipsia, polyuria, and polyphagia for 5 years; dysuria with poor glycemic control for 2 months." |
| Discharge Diagnoses | ["Type 2 Diabetes Mellitus", "Diabetic Ketosis", "Urinary Tract Infection", "Diabetic Macrovascular Complications", ...] |
| Discharge Medications | ["Acarbose", "Repaglinide", "Rosuvastatin", "Telmisartan", "Amlodipine", "Calcium Carbonate Tablets"] |

For model training and evaluation, the dataset was divided into training, validation, and test sets at a ratio of 6:1:3 based on unique patient identifiers. This ensured that multiple visits from the same patient were confined to a single subset, thereby preserving data independence and avoiding information leakage. Detailed statistics are presented in **Table 2**.

CDrugRed captures realistic disease progression patterns and comorbidity profiles of Chinese patients with metabolic diseases, offering a valuable benchmark for advancing data-driven clinical decision support and intelligent medication recommendation research in Chinese healthcare settings.

Table 2. Statistical summary of the CDrugRed dataset

| Dataset | Number of Patients | Number of Visits |
|---|---|---|
| Train | 1910 | 3602 |
| Validation | 320 | 570 |
| Test | 960 | 1722 |

### 3.3   Baseline System

To provide a reference for comparison with the participants' systems, we established an LLM-based baseline model. Specifically, we conducted supervised fine-tuning (SFT) based on the GLM4-9B-Chat model using the LoRA method. The fine-tuning hyperparameters are summarized in **Table 3**, and the corresponding instruction template is illustrated in **Fig. 2.** . For the instruction training data construction, all patient information fields excluding discharge medications were concatenated to form the instruction input, while the corresponding discharge medication list was used as the output. This instruction-response formulation enables the model to learn the mapping between clinical context and medication recommendation, thereby establishing a large language model–based baseline for this task.

Table 3. LoRA fine-tuning hyperparameters for GLM4-9B-Chat

| Hyperparameter | Value |
|---|---|
| Batch size | 1 |
| Gradient accumulation | 4 |
| Learning rate | 1e-4 |
| LoRA rank | 8 |
| LoRA alpha | 16 |
| Epoch | 10 |

| SFT Prompt Template (zh) | SFT Prompt Template (en) |
|---|---|
| 你作为三甲医院主任医师，需根据患者完整病历和药物列表，严格遵循临床规范生成出院带药方案。病历如下：<br>【患者核心信息】<br>• 基础体征：{sex}/{age}岁（注意年龄相关剂量调整）<br>• 过敏史：{allergy}<br>• 生理指标：身高{height}cm 体重{weight}kg BMI{bmi}<br>【临床诊疗关键点】<br>▶ 诊疗过程描述：{process}<br>▶ 入院情况：{admission}<br>▶ 主诉：{complaint}<br>▶ 现病史：{history_now}<br>▶ 既往史：{history_past}（关注慢性病用药史）<br>▶ 出院诊断：{diagnosis}（核心依据）<br>【待选药物库】{drug_str}<br>【输出规范】<br>请从给定的药物列表中推荐该患者的"出院带药列表"，请仅输出以下格式："药物A,药物B,药物C,......"<br>严格按照格式回答，不要输出任何解释或额外内容。 | You are a chief physician at a top-tier hospital. Based on the patient's complete medical record and medication list, you must strictly follow clinical guidelines to generate a discharge medication plan. The medical record is as follows:<br>[Patient Core Information]<br>• Basic vitals: {sex}/{age} years old (pay attention to age-related dose adjustments)<br>• Allergy history: {allergy}<br>• Physiological indicators: Height {height} cm, Weight {weight} kg, BMI {bmi}<br>[Key Clinical Points]<br>▶ Description of treatment process: {process}<br>▶ Admission condition: {admission}<br>▶ Chief complaint: {complaint}<br>▶ Present illness history: {history_now}<br>▶ Past medical history: {history_past} (pay attention to chronic medication history)<br>▶ Discharge diagnosis: {diagnosis} (core basis)<br>[Medication Library for Selection] {drug_str}<br>[Output Specification]<br>Please recommend this patient's "discharge medication list" from the given medication options.<br>Only output in the following format:<br>"DrugA, DrugB, DrugC, ..."<br>Answer strictly in this format. Do not output any explanation or additional content. |

**Fig. 2.** Instruction template used for baseline model training. (Note that the English prompt is not part of the input, it is the translation of the Chinese.)

### 3.4 Evaluation Metric

Since the discharge medications are restricted to a predefined set of 651 candidate drugs, this task can be treated as a multi-label classification problem. Accordingly, two commonly used metrics, i.e., Jaccard score and F1 score, are adopted to comprehensively evaluate system performance.

$$Jaccard = \frac{1}{N} \sum_{i=1}^{N} \frac{|y_i \cap \hat{y}_i|}{|y_i \cup \hat{y}_i|} \tag{1}$$

$$Precision(y_i, \hat{y}_i) = \frac{|y_i \cap \hat{y}_i|}{|\hat{y}_i|} \tag{2}$$

$$Recall(y_i, \hat{y}_i) = \frac{|y_i \cap \hat{y}_i|}{|y_i|} \tag{3}$$

$$f1(y_i, \hat{y}_i) = 2 \cdot \frac{Precision(y_i, \hat{y}_i) \cdot Recall(y_i, \hat{y}_i)}{Precision(y_i, \hat{y}_i) + Recall(y_i, \hat{y}_i)} \tag{4}$$

In the metrics, $y$ denotes the set of ground-truth drugs, $\hat{y}$ denotes the set of drugs predicted by the model, and $|X|$ denotes the number of elements in set $X$.

To obtain overall system performance, macro-averaged precision, recall, and F1 scores are computed across all $N$ patient records:

$$AVG\_P = \frac{1}{N} \sum_{i=1}^{N} Precision(y_i, \hat{y}_i) \tag{5}$$

$$AVG\_R = \frac{1}{N} \sum_{i=1}^{N} Recall(y_i, \hat{y}_i) \tag{6}$$

$$F1 = \frac{1}{N} \sum_{i=1}^{N} f1(y_i, \hat{y}_i) \tag{7}$$

Finally, the overall ranking score used for leaderboard evaluation is defined as the average of the Jaccard and F1 scores, which balances the model's performance in terms of both label overlap and predictive accuracy:

$$Score = \frac{1}{2} \cdot (Jaccard + F1) \tag{8}$$

This composite metric ensures a fair and comprehensive assessment of system performance across all participating teams.

## 4 Results

A total of 526 teams registered for this shared task. Among them, 167 teams submitted valid results to the Phase A leaderboard, and 95 teams submitted results to the Phase B leaderboard. These teams are from a diverse range of organizations, including universities, enterprises, and research institutions. Each team was allowed to submit up to three runs per day, and the highest historical score was recorded as the team's final leaderboard result.

### 4.1 Evaluation Results

During Phase A, participants were evaluated on the validation set, while Phase B used the test set for the final leaderboard ranking. The best results achieved by the participants, along with the baseline performances in both phases, are summarized in **Table 4**.

Table 4. Best results and baseline performances on both leaderboards

| Phase | Type | Jaccard | AVG_P | AVG_R | F1 | Score |
|---|---|---|---|---|---|---|
| Leaderboard A | Best result | 0.5025 | 0.6764 | 0.6077 | 0.6164 | 0.5594 |
| Leaderboard A | Baseline | 0.4444 | 0.5751 | 0.5958 | 0.5621 | 0.5032 |
| Leaderboard B | Best result | 0.5102 | 0.6905 | 0.6121 | 0.6267 | 0.5685 |
| Leaderboard B | Baseline | 0.4477 | 0.5864 | 0.5872 | 0.5648 | 0.5062 |

As shown in Table 4, the top-performing systems achieved notable improvements over the baseline, particularly in the Jaccard and F1 scores. The best-performing model in Phase A and B achieved final scores of 0.5594 and an F1 score of 0.5685, outperforming the baseline by 5.62% and 6.23%, respectively.

**Table 5** further presents the results of the top ten teams on Leaderboard B. The first-ranked team, DeepDrug, achieved the highest overall score of 0.5685, establishing a clear lead over other teams. In contrast, the performance gap among the second to tenth teams is relatively small, with overall scores ranging from 0.5226 to 0.5453, suggesting a high degree of competitiveness in this range.

Table 5. The performance results of top 10 ranking teams.

| Team | Rank | Jaccard | AVG_P | AVG_R | F1 | Score |
|---|---|---|---|---|---|---|
| DeepDrug | 1 | 0.5102 | 0.6905 | 0.6121 | 0.6267 | 0.5685 |
| ZZUNLP | 2 | 0.4876 | 0.6897 | 0.5735 | 0.6031 | 0.5453 |
| suxiao818 | 3 | 0.4870 | 0.6740 | 0.5810 | 0.6014 | 0.5442 |
| 晚安东莞 | 4 | 0.4790 | 0.6333 | 0.6070 | 0.5965 | 0.5378 |
| H-3-C | 5 | 0.4732 | 0.6430 | 0.5857 | 0.5899 | 0.5316 |
| 对对队 | 6 | 0.4721 | 0.6194 | 0.6019 | 0.5883 | 0.5302 |
| Med-LLM | 7 | 0.4697 | 0.6317 | 0.5812 | 0.5851 | 0.5274 |
| IIGROUP | 8 | 0.4669 | 0.6345 | 0.5858 | 0.5850 | 0.5260 |
| Seeking Your Roots | 9 | 0.4644 | 0.6068 | 0.6036 | 0.5826 | 0.5235 |
| 熬夜美少女战士 | 10 | 0.4628 | 0.6276 | 0.5771 | 0.5824 | 0.5226 |
| Baseline | 28 | 0.4477 | 0.5864 | 0.5872 | 0.5648 | 0.5062 |

Overall, 27 teams outperformed the baseline on Leaderboard B. These results show that many participating teams effectively leveraged model fine-tuning, data augmentation, and ensemble strategies to substantially enhance predictive performance compared to the baseline. The consistent improvement across multiple teams underscores the potential of LLMs for clinical decision support tasks, particularly when combined with domain-specific data augmentation and prompt design strategies.

### 4.2 Descriptions of Top Five Teams

**Team DeepDrug (Tencent Jarvis Lab)**

The DeepDrug team proposed a generative recommendation framework integrating multi-dimensional feature enhancement and multi-scale model fusion. At the data representation level, they introduced drug category annotations, explicit patient meta-features, and disease–drug co-occurrence knowledge to enhance clinical semantic understanding. They also applied order perturbations to diagnostic and medication lists to improve model robustness and generalization. At the model level, they performed full supervised fine-tuning based on the Qwen-series LLMs and adopted differentiated training strategies for models of different parameter sizes to obtain complementary ensembles. During inference, a hierarchical weighted-voting fusion mechanism was introduced, assigning adaptive weights based on validation performance to balance bias and variance across model predictions.

**Team ZZUNLP (Zhengzhou University)**

The ZZUNLP team developed a systematic solution across the dimensions of data, model training, and inference. To address the domain specificity of medical text and the scarcity of labeled data, they designed prompt templates tailored for clinical medication reasoning and introduced a pseudo-labeling data-augmentation strategy. They used the cross-entropy loss of label sequences as a confidence indicator to filter high-quality pseudo-labeled samples. Multiple open-source LLMs were fine-tuned using LoRA and optimized with the liger-kernel cross-entropy loss for efficient memory utilization. For inference, they employed a two-stage hierarchical ensemble: in stage one, weighted voting was applied separately within model groups trained on original data and augmented data to obtain high-confidence predictions; in stage two, an adaptive fusion strategy based on differences in predicted list length and Jaccard similarity further improved final recommendation performance.

**Team suxiao818 (MISUMI (China) Precision Machinery Trading Co., Ltd.)**

The suxiao818 team implemented a fine-tuning and ensemble-based framework built upon multiple LLMs. They split the dataset (9:1) into training and validation subsets and selected the top-performing four models with LoRA fine-tuning for ensemble. To mitigate randomness in generative decoding, each model produced multiple predictions under different decoding temperatures, which were fused using majority voting. Candidate drugs appearing in more than half of the predictions were retained and ranked by frequency, with list length truncated to the average.

**Team 晚安东莞 (South China Normal University)**

This team fine-tuned Qwen2.5-7B-Instruct for discharge medication generation. In data preprocessing, they designed prompt strategies for electronic medical records, concatenating all fields of a case as model input to generate structured or natural-language-style drug lists. They then performed supervised fine-tuning via LoRA. At inference, they ran multiple temperature settings for models outputting the different styles, and selected the three best-performing configurations for ensemble. They further normalized drug names to improve consistency. The final ensemble used a voting rule: drugs appearing in at least two lists were retained; if there was no overlap, the union of all predictions was used to increase recall.

**Team H-3-C (H3C Technologies Co., Limited)**

The H-3-C team designed a data-centric approach emphasizing rare-drug oversampling and ensemble enhancement. Low-relevance fields (such as patient index, gender, BMI) were removed, and samples containing rare drugs were duplicated or triplicated based on frequency thresholds. To increase data diversity, they introduced order-shuffled samples and combined them with the oversampled set, resulting in 6,740 augmented samples. Five open-source LLMs were fine-tuned using LoRA on these datasets. During inference, fuzzy string matching normalized predicted drugs to the candidate list, supplemented by a manually curated alias mapping table. Final predictions were generated via majority voting (appearing in ≥3 models), with sequential back-filling from GLM4-9B-Chat and CareBot_Medical_multi-llama3-8b-instruct models for empty outputs.

### 4.3  Analysis of Top Team Solutions

**Table 6** presents the performance of the top five teams on the Leaderboard B. Since all these teams adopted model ensemble strategies, we report both their best single-model (Single) results and their best ensembled results. The ensemble approaches generally led to consistent performance gains across all teams, highlighting the importance of model diversity and fusion in this multi-label clinical prediction task.

Table 6. Single and ensemble results of the top five ranking teams on the test set. Single refers to Best Single Model

| Team | Rank | Method | Jaccard | AVG_P | AVG_R | F1 | Score |
| --- | --- | --- | --- | --- | --- | --- | --- |
| DeepDrug | 1 | Single | - | - | - | - | - |
|  |  | Ensemble | 0.5102 | 0.6905 | 0.6121 | 0.6267 | 0.5685 |
| ZZUNLP | 2 | Single | 0.4480 | 0.5837 | 0.5902 | 0.5648 | 0.5064 |
|  |  | Ensemble | 0.4876 | 0.6897 | 0.5735 | 0.6031 | 0.5453 |
| suxiao818 | 3 | Single | 0.4700 | 0.6498 | 0.5750 | 0.5865 | 0.5283 |
|  |  | Ensemble | 0.4870 | 0.6740 | 0.5810 | 0.6014 | 0.5442 |
| 晚安东莞 | 4 | Single | 0.4644 | 0.6044 | 0.6082 | 0.5821 | 0.5232 |
|  |  | Ensemble | 0.4790 | 0.6333 | 0.6070 | 0.5965 | 0.5378 |
| H-3-C | 5 | Single | 0.4538 | 0.5965 | 0.5946 | 0.5715 | 0.5127 |
|  |  | Ensemble | 0.4732 | 0.6430 | 0.5857 | 0.5899 | 0.5316 |

Notably, the second-place team (ZZUNLP) achieved the largest performance gain after ensemble fusion, with its score increasing from 0.5064 to 0.5453, underscoring the effectiveness and sophistication of its two-stage ensemble strategy. This approach combines intra-group weighted voting with inter-group adaptive fusion. It demonstrates that carefully designed hierarchical ensemble mechanisms can capture complementary model strengths more effectively than simple averaging or majority voting. Overall, these findings highlight that, while large language models provide strong baselines, ensemble-based optimization remains a crucial technique for improving the stability and accuracy of clinical drug recommendation systems.

## 5    Conclusion

In this paper, we established the first benchmark for discharge medication recommendation in metabolic diseases using Chinese EHR data through the CHIP 2025 Shared Task 2. The high level of participation and diverse system designs reflect strong community engagement and the growing interest in data-driven clinical decision support. The top-performing teams demonstrated that ensemble frameworks combining domain-specific fine-tuned LLMs can effectively leverage EHR information to improve discharge medication prediction. This shared task provides valuable insights into the practical application of NLP and machine learning for personalized medication management, contributing to the advancement of precision medicine in chronic metabolic care.

Despite these advances, challenges remain in addressing label imbalance, rare medication usage, and generalizing across institutions. For future work, we plan to expand the dataset with additional clinical contexts from multiple institutions, incorporating multi-modal data such as laboratory and medical imaging, and moving beyond drug name recommendation to generating complete medication regimens, including dosage and instructions. These efforts aim to promote explainable and trustworthy AI systems for clinical medication applications.

**Acknowledgments.** This research was supported by the Natural Science Foundation of China (No. 62302076, 62276043), the Fundamental Research Funds for the Central Universities (No. DUT25YG108), and the Research Project on High Quality Development of Hospital Pharmacy, National Institute of Hospital Administration, NHC, China (No. NIHAYSZX2525).

Overview of CHIP 2025 Shared Task 2     13